\documentclass[10pt,twocolumn,letterpaper]{article}

\usepackage{cvpr}
\usepackage{times}
\usepackage{epsfig}
\usepackage{graphicx}
\usepackage{amsmath}
\usepackage{amssymb}
\usepackage{pifont}
\newcommand{\cmark}{\ding{51}}%
\usepackage{booktabs} 
\usepackage{color}

\usepackage[T1]{fontenc}
\usepackage[utf8]{inputenc}
\usepackage{authblk}
\usepackage[misc]{ifsym}

\usepackage[breaklinks=true,bookmarks=false]{hyperref}

\cvprfinalcopy 

\def\cvprPaperID{****} 

\ifcvprfinal\fi
\begin{document}

\title{IPG-Net: Image Pyramid Guidance Network for Small Object Detection}

\author[1]{Ziming Liu}
\author[1]{Guangyu Gao\thanks{corresponding author}}
\author[2]{Lin Sun}
\author[1]{Li Fang}

\affil[1]{Beijing Institute of Technology, Beijing, China \authorcr {\tt\small liuziming.email@gmail.com}, {\tt\small  \{guangyugao,3220190791\}@bit.edu.cn}}
\affil[2]{Samsung Strategy and Innovation Center, California, US \authorcr {\tt\small lin1.sun@samsung.com}}

\maketitle
\thispagestyle{empty}

\begin{abstract}

For Convolutional Neural Network-based object detection, there is a typical dilemma: the spatial information is well kept in the shallow layers which unfortunately do not have enough semantic information, while the deep layers have a high semantic concept but lost a lot of spatial information, resulting in serious information imbalance. To acquire enough semantic information for shallow layers, Feature Pyramid Networks (FPN) is used to build a top-down propagated path. In this paper, except for top-down combining of information for shallow layers, we propose a novel network called Image Pyramid Guidance Network (IPG-Net) to make sure both the spatial information and semantic information are abundant for each layer. Our IPG-Net has two main parts: the image pyramid guidance transformation module and the image pyramid guidance fusion module. Our main idea is to introduce the image pyramid guidance into the backbone stream to solve the information imbalance problem, which alleviates the vanishment of the small object features. This IPG transformation module promises even in the deepest stage of the backbone, there is enough spatial information for bounding box regression and classification. Furthermore, we designed an effective fusion module to fuse the features from the image pyramid and features from the backbone stream. We have tried to apply this novel network to both one-stage and two-stage detection models, state of the art results are obtained on the most popular benchmark data sets, i.e. MS COCO and Pascal VOC.

\end{abstract}

\section{Introduction}
Recently, with the development of deep convolution neural networks, there have been abundant CNN based methods focusing on object detection tasks since the emergence of typical networks of Faster-RCNN \cite{fasterrcnn}, YOLO \cite{redmon2018yolov3}, SSD \cite{SSD}, RetinaNet \cite{retina} etc. However, object detection still suffers from some problems, such as the key problem of information imbalance of different feature scales. Because the convolution neural network is designed to output a single output for classification, not for the multi-scale tasks. 

Some works have tried to fix this imbalance, such as the most popular Feature Pyramid Network (FPN), which mainly fixed the problem of lacking high semantic information in shallow layers. 

Although feature pyramid network can supply the semantic information for shallow features, there are still feature misalignment and information lost in deeper features, which is especially harmful for small object detection. Feature misalignment refers to that there are some offsets between anchors and convolution features.

\begin{figure*}[t]
    \includegraphics[width= \textwidth]{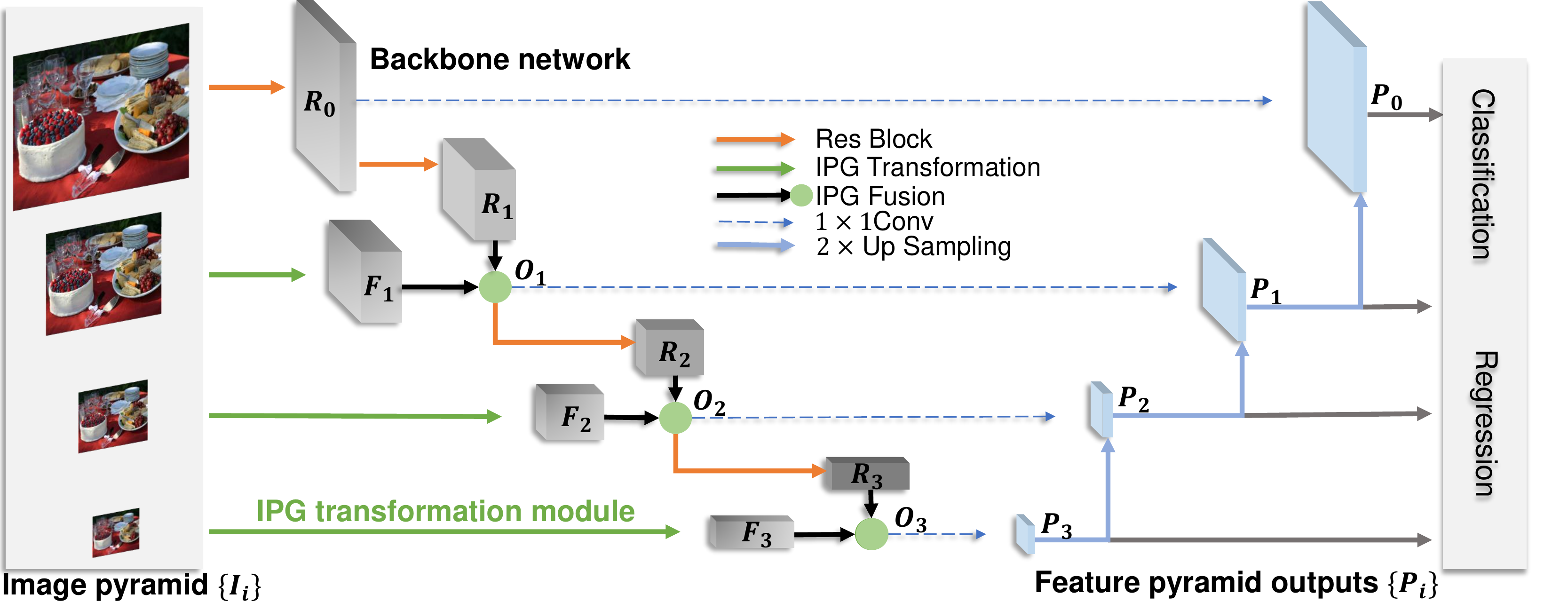}
    \caption{The overall structure of the IPG-Net, including two main parts: IPG transformation module and the IPG fusion module. The input is an image pyramid $\{I_{i}\}$. The green part (IPG transformation module and IPG fusion module) is what proposed in this paper. The blue part is the multi-scale feature pyramid outputs $\{P_{i}\}$. }
    \label{fig:backbone}
\end{figure*}

In this paper, we argue that good feature extractor for detection should have two common features: i) enough shallow image information for bounding box regression because object detection is a typical regression task. ii) enough semantic information for classification, which means the output features come from deep layers. To satisfy these characters above, we introduce a novel network specific for object detection, namely, the Image Pyramid Guidance Network (IPG-Net). The IPG-Net includes two main parts: the \textit{IPG transformation module and the IPG fusion module}, as shown in Fig. \ref{fig:backbone}. The IPG-Net is designed for extracting better features by fixing the information imbalance problem better.

The deep convolution network will cause the loss of the location or spatial information as the layer becomes deeper. This property maybe not a problem for the classification task, while bounding box regression is important for the detection task. But, the loss of such spatial information results in the features misalignment in object detection. Here, feature misalignment means there are some offsets between anchors and convolution features. Besides the loss of spatial information, small objects will easy to be lost in the deeper convolution layers. We argue that all these problems for object detection are due to the limit of the existed convolution network structure and can't be fixed by just simply modifying typical networks' architecture.

Here, we introduce the image pyramid to supply more spatial information into each stage of the feature pyramid of the backbone network. Then the mentioned problems can be reduced in this way. For each stage of the backbone network, we compute the image pyramid feature of the corresponding level in the image pyramid. The image pyramid feature is obtained from a shallow and light-weighted IPG transformation module, which has more abundant spatial information, especially for small objects, compared with the deep backbone. Then we design an IPG fusion module to fuse the new image pyramid feature into the backbone network.

The fusion module performs two steps to fuse the two kinds of features. Firstly, we transform the original features to align the data size and project them into a hidden space. Secondly, We use common mathematics operations to combine the two features. Sum, product, and concatenation are all used in our experiments and improvements of different degrees are obtained.

Before going deeper into our proposed methods, we summarize our contributions as below:
\begin{itemize}
    \item We introduce the image pyramid guidance (IPG) into the backbone stream(network) to fix the spatial information and small objects' features lost problem in deep layers.
    \item We design a new shallow IPG transformation module to extract image pyramid features, which is flexible and light-weighted.
    \item We also design a flexible fusion module, which is simple but effective.  
\end{itemize}

\section{Related Work}
Object detection is a basic task for deeper visual reasoning or visual understanding. The state-of-the-art works based on deep learning for object detection can be classified into one stage model and two-stage model(Faster RCNN\cite{fasterrcnn}, Cascade RCNN\cite{cai2018cascade}, SNIP\cite{snip},SNIPER\cite{singh2018sniper} etc.), and one stage model can be further be classified into anchor-based methods(Retina net\cite{retina}, Yolo-v3\cite{redmon2018yolov3} etc.) and anchor free methods(Center net\cite{duan2019centernet}, FSAF\cite{fsaf} etc.). All of SOTA models are based on the 3 branches, two-stage methods are easier to achieve slightly better results while one stage methods have faster speed in practice. There are also some works about design backbone network specific for object detection as what we do here, Detnet is some of them\cite{li2018detnet}.

\subsubsection{Two stage detector}
Two-stage algorithms keep the state of the art results in most popular data sets, such as MS COCO\cite{coco}, Pascal VOC\cite{pascalvoc}. However, they also suffer from the speed limit and the huge complexity of the model building. The information imbalance is also a tough problem for two-stage algorithms, although there are some works reduce the imbalance impact to some degree, such as feature pyramid network\cite{fpn}, this is still an unsolved problem.

\subsubsection{One stage detector}
To achieve faster inference speed, a lot of one stage algorithms were proposed and achieved as good performance as two-stage models. The initial SOTA one stage models are based on anchor mechanism, but more efficient algorithms of anchor free are proposed recently. The most typical works including center net which motivated by key point detection\cite{duan2019centernet}, WSMA-Seg which is motivated by segmentation\cite{cheng2019segmentation}, FSAF\cite{fsaf}. Unfortunately, the information imbalance and the feature misalignment also impact the one-stage methods' performance, especially the anchor-based detectors.


\subsubsection{Information imbalance and Feature alignment}
There are also some works to solve the imbalance problem at the feature level. PANet \cite{panet} added a bottom-up path on previous FPN to shorten the information propagate path between lower features and the topmost feature. Pang etc. proposed Libra R-CNN which contains a balanced feature pyramid to reduce the imbalance in feature level, i.e. the outputs of the feature pyramid network(FPN) \cite{pang2019libra}. EFIP \cite{EFIPPang_2019_CVPR} introduced an independent network to extract features from images of different resolution, and then fuse these features with the standard SSD \cite{SSD} outputs. Although they use an image pyramid as input, they only modify the final output layer of SSD \cite{SSD}. As we discussed above, information imbalance and misalignment problems happen inside the backbone network. To solve that, we let IPG-Net continually fuse the image pyramid information into the backbone stream. All of the works above are trying to fix the imbalance and misalignment problem, but there is still no one that can solve the problem completely in object detection. Here we propose a novel network, IPG-Net, which is based on an image pyramid. Fusing the image pyramid into the detection backbone to solve the information imbalance problem is a new path.

\section{Image Pyramid Guidance Network(IPG-Net)}

\subsection{Challenges to be Solved}
FPN reduces the information imbalance of features of different scales to some degree, but there are still challenges waiting to be solved. we summary these challenges in this section.
\subsubsection{Anchor Misalignment.} 
Although deeper CNN enables better semantic features to be extracted, it also blurs these features. The location of objects in deep features is not always aligned with the location of those objects in original images. But anchor-based detection algorithms follow the assumption that the location of objects in any feature is aligned with that in corresponding original images \cite{retina,fasterrcnn,yolov2,redmon2018yolov3}. Therefore, there is a serious misalignment between the anchor and the convolution features. This phenomenon becomes more serious with the increase of CNNs depth \cite{pang2019libra}.
\subsubsection{FPN Misalignment.}
Feature pyramid network fuses deep features to the corresponding shallow features to alleviate the information imbalance problem. However, because deep CNN backbone already causes anchor misalignment in deep features, The fusing of FPN can't make the right alignment between deep features and corresponding shallow features. For example, without image pyramid guidance, because there is already misalignment problem between feature $R_{2}$ and feature $R_{1}$ as mentioned in the last section, the feature $P_{1}=upsample(P_{2})+Conv(R_{1})$ will also suffer the misalignment problem.


\subsubsection{Feature Vanishment for Small Objects.}
Deep CNNs achieve high performance in classification due to the large stride of 32 respecting to initial image size. However, large stride also leads to the miss of the detailed information of the input image, i.e. the small object information. Small object detection depends on detailed information. Therefore, we usually detect small objects with shallow layer's features. But these features lack semantic information. Using feature pyramid network (FPN) to build a top-down path to supply semantic information for shallow layers' features is essential. Although FPN improves the detection difficulty in shallow layers to some degree, there is still a serious loss of those small object information. Because this detail information of small/tiny objects has been largely damaged in the deep layer of CNN backbone. This is also why we propose to supply shallow layer information to deep layer with image pyramid guidance (IPG). 

\subsection{Overall Structure}

The overall structure of the image pyramid guidance net (IPG-Net) is shown in Fig. \ref{fig:backbone}. IPG-Net is modified from the traditional backbone network, such as ResNet \cite{resnet}, which could provide a fair comparison with the existing methods. There are two main parts in IPG-Net: \textit{IPG transformation module, IPG fusion module}.

The IPG transformation module accepts a set of images of different resolutions from the image pyramid and extracts the image pyramid features for fusing. The function of the IPG transformation module is to extract shallow features to supply spatial information and detail information. The image pyramid features are used to guide the backbone network to reserve spatial information and small objects' features. Furthermore, We use a fusion module to perform the guidance. The IPG fusion module's function is to fuse the deep features of the backbone network and the shallow features of IPG transformation module, the formulation and variants will be discussed in the next section. The idea of the IPG fusion module is to make a transformation of the two types of features firstly and then fuse them together to achieve an augment effect for the object detection, especially small/tiny object detection. 

\subsection{IPG Transformation Module}
\begin{figure}[t]
    \centering
    \includegraphics[width=8cm]{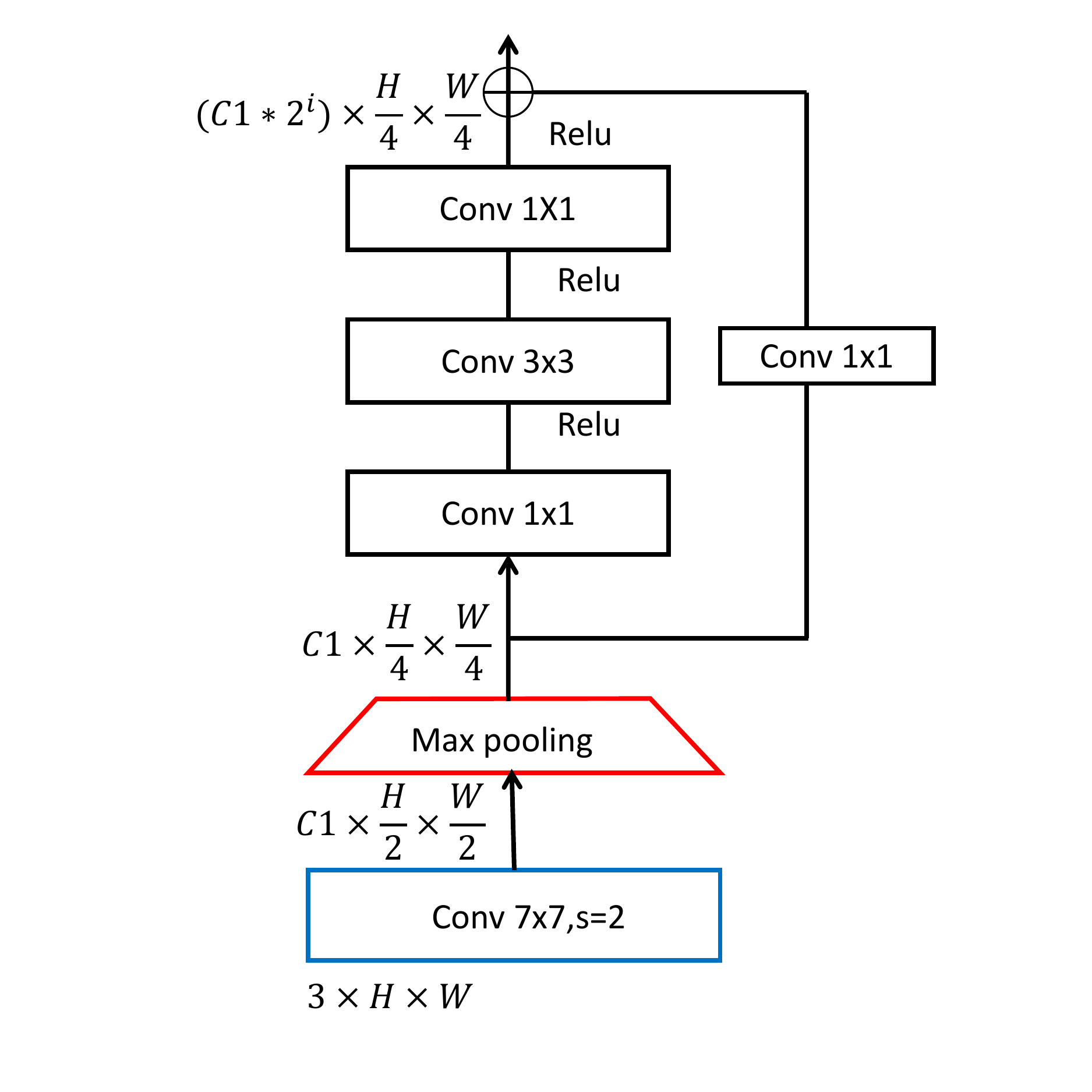}
    \caption{The structure of the IPG transformation module of level $i$, $i$ ranges from $1$ to $N-1$, $N$ is the total depth(stages) of the backbone. The IPG transformation module has different parameters at different levels. The output feature's channel dimension of level $i$ is $(C1\times 2^{i})$, $C1$ is usually $256$ if using a ResNet50 backbone. This channel dimension is consistent with the backbone feature.}
    \label{fig:subnet}
\end{figure}

Traditionally, an image pyramid is used to obtain multi scales feature to reduce the influence of image scale, because the CNN lacks the scale-invariant ability. Usually, The performance of most models can be significantly improved in this way, but the computation cost is also too large to afford in the training stage, especially for a deep CNN. Different from the traditional method, in this paper, we use the image pyramid to guide the backbone network to reduce the information imbalance problem and learn better detection features. Better features mean that these features of different scales have both abundant spatial information and enough semantic information, i.e. there are no serious feature misalignment and information imbalance.

The input of the IPG transformation module is an image pyramid set $IPset = \{I_{i}\}, i\in [0,N)$. The image resolutions of images in $IPset$ decrease with $2$ times. The first image is $I_{0}$ with $H\times W$ resolution which is the same as the commonly used image resolution of object detection. $N$ is the number of levels of the image pyramid. We set $N=4$ in our experiments to be consistent with the depth of a standard ResNet. 

Next, we will introduce the typical structure of the IPG transformation module, as shown in Figure \ref{fig:subnet}. The structure of IPG transformation module is component with two parts, one is a $7\times7$ convolution followed with a $2\times2$ max pooling, another is a residual block, which is similar to the residual design in \cite{resnet}. The residual block accepts features of the same dimensions but outputs features of different dimensions, the output dimension of features are aligned with that of the backbone network. There are two main reasons why we use a shallow sub-network to extract image pyramid feature. On the one hand, the shallow layer could reserve more spatial information/detail information, while deep CNN will damage this information. On the other hand, the computation cost and the number of network parameters will not increase too much because of the shallow and light-weighted design.

Each component of the outputs of the IPG transformation modules $IPFset = \{F_{i}\}, i\in [0,N)$ can be formulated as: $F_{i} = f(I_{i}), i\in [0,N)$.
where the $f( \cdot )$ denotes the IPG transformation module, as shown in Figure \ref{fig:subnet}, $F_{i}$ denotes the image pyramid feature of the level $i$. Those features $F_{i}$ from different level of image pyramid $IPset = \{I_{i}\}, i\in [0,N)$ form new image pyramid features set $IPFset = \{F_{i}\}, i\in [0,N)$ . 

\begin{figure}[t]
    \centering
    \includegraphics[width=8cm]{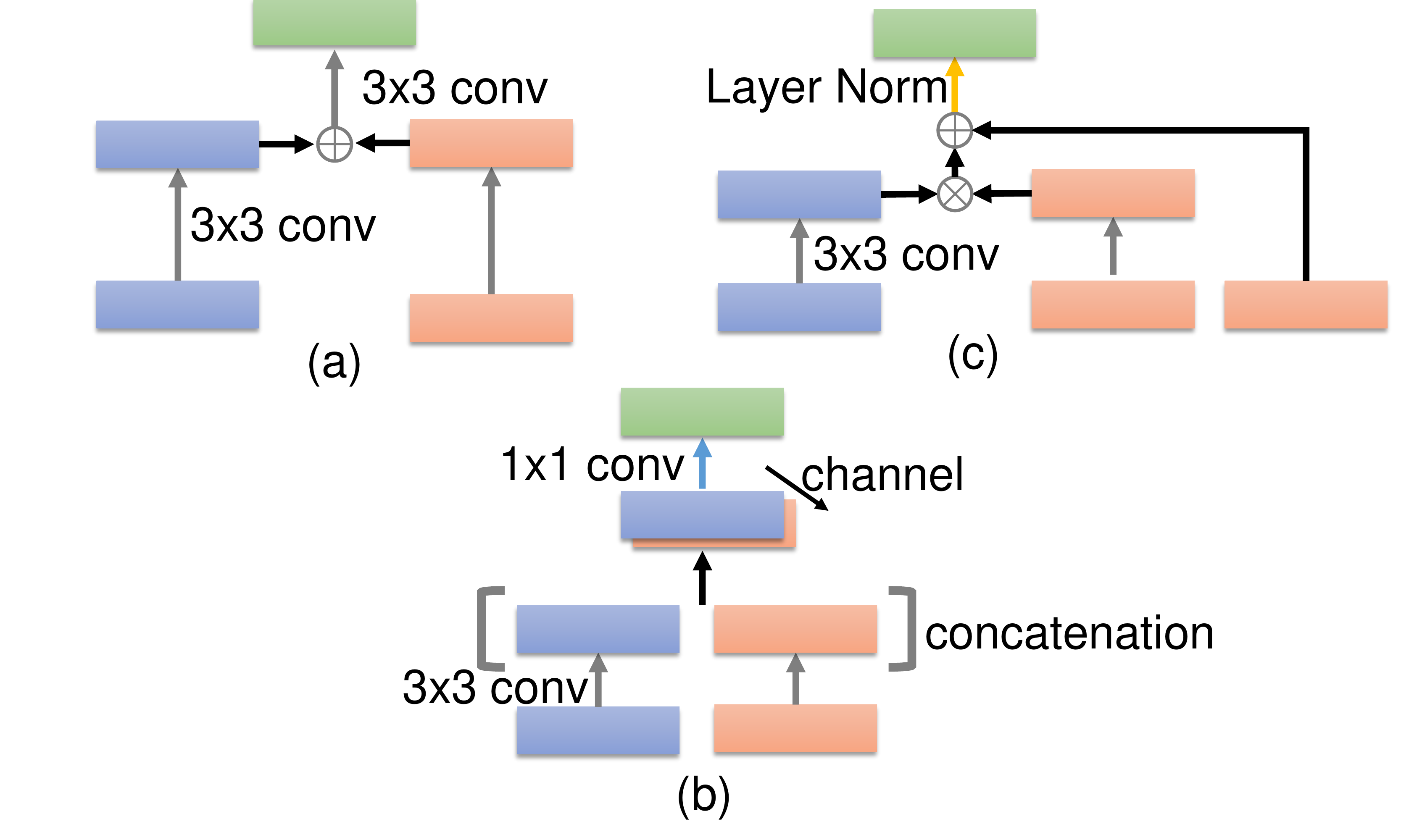}
    \caption{Three instances of the fusion module, $(a)$ is the sum-up strategy, $(b)$ is the concatenation strategy, $(c)$ is the residual product strategy. All of them are the variants of the Eq. \ref{eq:fusing}.}
    \label{fig:fusingmodule}
\end{figure}

\subsection{Backbone Network}
The backbone network of IPG-Net is modified from the standard ResNet which contains four stages (stage1-stage4). In this paper, We add new stages at the end of standard ResNet, each new stage contains two Bottleneck modules, same as ResNet. Our ablation studies suggest adding one new stage can perform better than the other conditions. Too deep backbone network also is harmful for the detection, We argue that the backbone which is too deep has difficulty for training, similar to the classification task.

The reason why we use a deeper convolution network than the standard ResNet is that the IPG transformation module supplies enough spatial information/detail information into the backbone network, which promises we could train a deep CNN without much information imbalance or misalignment. A deeper backbone network enables us to generate better semantic information which is good for the classification and could cover a larger range of object scales.

\subsection{IPG Fusion Module}

\subsubsection{Formulation}
The IPG fusion module in this paper is a flexible module, we first formulate it as follows. The $f(\cdot)$ and $g(\cdot)$ correspond to the network of IPG transformation module and backbone network separately. The function of $\beta$ can be flexible with different versions.
\begin{equation}
    O_{i} = \beta(f_{i}(I_{i}),g_{i}(I_{0})), \\i \in [1,N-1]
    \label{eq:fusing}
\end{equation}
where $O_{i}$ is the output feature the of fusion module in level $i$, as shown in Figure \ref{fig:backbone}. $I_{0}$ and $I_{i}$ are images in the image pyramid in level $0$ and level $i$ separately. The $\beta(\cdot)$ denotes the basic fusion function of the fusion module. The $f_{i}(\cdot)$ denotes the output of the IPG transformation module in level $i$ and the $g_{i}(\cdot)$ denotes the output $R_{i}$ of the backbone network in level $i$. The number of levels $N$ is determined by the size of image pyramid $IPset$.

The position of IPG fusion module in IPG-Net is shown in Figure \ref{fig:backbone}. For each IPG fusion module, There are two inputs, the image pyramid feature $F_{i}$ and the corresponding backbone feature $R_{i}$. Further, We propose several different variants of IPG fusion module to demonstrate the effectiveness of image pyramid guidance. Sum, Product, and Concatenation are the three types of fusion modules we used in our experiments. The other similar design of the fusion module will also work well, such as the attention-based design, but we will not focus on that in this paper. we will follow this idea in our future work. 

Next, we will describe the details of three types of variants.



\subsubsection{Element-wise Sum}
For this version, we regard the image pyramid information as additional information. Therefore, the goal is to sum the image pyramid feature $F_{i}$ and backbone feature $R_{i}$ together. Firstly, we need to align the channel dimension of these two types of features. Here, we use channel-dimension linear interpolate operation to perform the $CT(channel transform)$. 
\begin{equation}
    O_{i} = W\cdot [W_{s}\cdot CT(F_{i})) + W_{m}\cdot R_{i})]
\end{equation}
Where the $W, W_{s}, W_{m}$ denotes different linear transformations. 
\subsubsection{Residual Product}
Here we use the dot product $W_{s}\cdot CT(F_{i})) * W_{m}\cdot R_{i}$ to represent the lost information in backbone feature $R_{i}$. After adding the lost information into backbone feature, a "layer norm" operation is performed to normalize the fused feature $O_{i}$.

\begin{equation}
    O_{i} =  LN \{ [W_{s}\cdot CT(F_{i})) * W_{m}\cdot R_{i}] + R_{i}) \}
\end{equation}

Where the $LN$ denotes the Layer Norm operation.

\subsubsection{Concatenation}
We also try to use concatenation operation to realize the fusing of the image pyramid feature and the backbone feature, which is similar to the fusing operation in U-net\cite{unet}. The formulation is shown as following.
\begin{equation}
    O_{i} = W\cdot Cat[W_{s}\cdot CT(F_{i}) , W_{m}\cdot R_{i}]
\end{equation}
Where the $Cat$ denotes the concatenation operation.

\begin{table*}[t]
    \centering
    \setlength{\tabcolsep}{3.mm}{
    \begin{tabular}{c|c|c|c|c|c|c|c}
        \toprule
            \quad model&backbone&$AP$&$AP50$&$AP75$&$AP_{S}$&$AP_{M}$&$AP_{L}$\\
            \midrule
            \textit{Two Stage Det}\\
            \midrule
            R-FCN\cite{dai2016rfcn}&ResNet-101&29.9&51.9&-&10.8&32.8&45.0\\
            Faster RCNN++\cite{resnet}&ResNet-101&34.9&55.7&37.4&15.6&38.7&50.9\\
            Faster RCNN w FPN\cite{fpn}&ResNet-101&36.2&59.1&39.0&18.2&39.0&48.2\\
            DeNet-101(wide)\cite{denet}&ResNet-101&33.8&53.4&36.1&12.3&36.1&50.8\\
            CoupleNet\cite{couplenet}&ResNet-101&34.4&54.8&37.2&13.4&38.1&50.8\\
            Deformable R-FCN\cite{dai2016rfcn}&Aligned-Inception-ResNet&37.5&58.0&40.8&19.4&40.1&52.5\\
            Mask-RCNN\cite{maskrcnn}&ResNeXt-101&39.8&62.3&43.4&22.1&43.2&51.2\\
            Cascade RCNN\cite{cai2018cascade}&ResNet-101&42.8&62.1&46.3&23.7&45.5&55.2\\
            SNIP++\cite{snip}&ResNet-101&43.4&\textbf{65.5}&48.4&\textbf{27.2}&46.5&54.9\\
            SNIPER(2scale)\cite{singh2018sniper}&ResNet-101&43.3&63.7&48.6&27.1&44.7&56.1\\
            Grid-RCNN\cite{gridrcnn}&ResNeXt-101&43.2&63.0&46.6&25.1&46.5&55.2\\
            
            \midrule
            \textit{Anchor based One Stage Det}\\
            \midrule
            SSD512\cite{SSD}&VGG-16&28.8&48.5&30.3&10.9&31.8&43.5\\
            YOLOv2\cite{yolov2}&DarkNet-19&21.6&44.0&19.2&5.0&22.4&35.5\\
            DSSD513\cite{fu2017dssd}&ResNet-101&31.2&50.4&33.3&10.2&34.5&49.8\\
            RetinaNet80++\cite{retina}&ResNet-101&39.1&59.1&42.3&21.8&42.7&50.2\\
            RefineDet512\cite{refinedet}&ResNet-101&36.4&57.5&39.5&16.6&39.9&51.4\\
            M2Det800&VGG-16&41.0&59.7&45.0&22.1&46.5&53.8\\
            \midrule
            \textit{Anchor Free One Stage Det}\\
            \midrule
            CornetNet511\cite{cornernet}&Hourglass-104&40.5&56.5&43.1&19.4&42.7&53.9\\
            FCOS\cite{tian2019fcos}&ResNeXt-101&42.1&62.1&45.2&25.6&44.9&52.0\\
            FSAF\cite{fsaf}&ResNeXt-101&42.9&63.8&46.3&26.6&46.2&52.7\\
            CenterNet511\cite{duan2019centernet}&Hourglass-104&44.9&62.4&48.1&25.6&47.4&57.4\\
            \midrule
            IPG RCNN&IPG-Net101&\textbf{45.7}&64.3&\textbf{49.9}&26.6&\textbf{48.6}&\textbf{58.3}\\
        \bottomrule
    \end{tabular}}
    \caption{The state of the art of the performance on the MS COCO \textit{test-dev}, '++' denotes that the inference is performed with multi-scales etc.}
    \label{tab:voc}
\end{table*}

\section{Experiments}

\subsection{Experiment Details}

\textbf{Datasets.}
We conduct ablation experiments on two data sets, MSCOCO\cite{coco} and Pascal VOC\cite{pascalvoc}. MSCOCO is the most common benchmark for object detection, the COCO data set is divided into train, validation, including more than 200,000 images and 80 object categories. Following common practice, we train on the COCO \textit{train2017}(i.e. \textit{trainval 35k} in 2014) and test on the COCO \textit{val 2017} data set(i.e. \textit{minival} in 2014) to conduct ablation studies. Finally, we also report our state of the art results in MS COCO \textit{test-dev}, the test is finished in CodaLab\footnote{https://competitions.codalab.org/competitions/20794} platform. 

We also apply our algorithm on another popular data set, Pascal VOC. Pascal VOC 2007 has 20 classes and 9,963 images containing 24,640 annotated objects and Pascal VOC 2012 also has 20 classes and 11,530 images containing 27,450 annotated objects and 6,929 segmentation. We train our model with Pascal VOC 2007 \textit{trainval} set and Pascal VOC 2012 \textit{trainval} set and test the model with Pascal VOC2007 \textit{test}.

\textbf{Training.}
We follow the common training strategies for object detection, 12 epoch with 4 mini-batch in each GPU. All of the experiments are conducted in 8 NVIDIA P100 GPUs, optimized by SGD(stochastic gradient descent) and default parameters of SGD in pytorch framework are adopted. The learning rate is set as 0.01 at the beginning and decrease by a factor of 0.1 in epoch 7 and epoch 11. The linear warm-up strategy is also used, the number of warm-up iterations is 500 and the warm-up ratio is 1.0/3. All of the input images are resized into $1333\times800$ in COCO and $1000\times800$ in Pascal VOC, which is consistent with the common practice. The image pyramid is obtained by down-sampling(linear interpolate) the input image into four levels with a factor of 2.
\begin{table}[t]
    \centering
    \setlength{\tabcolsep}{2.7mm}{
    \begin{tabular}{c|c|c|c|c}
        \toprule
            \quad model&fusing strategy&$AP_{S}$&$AP_{M}$&$AP_{L}$\\
            \midrule
            IPG RCNN&sum&\textbf{20.8}&\textbf{39.6}&\textbf{46.2}\\
            &product&\textbf{18.9}&36.3&43.4\\
            &concatenation&\textbf{19}&35.5&42.6\\
        \bottomrule
    \end{tabular}}
    \caption{The ablation study of the fusion module on the MS COCO \textit{minival}.}
    \label{tab:fusingstrategy}
\end{table}

\textbf{Inference.}
The image size of the image pyramid keeps the same with the training stage. The IOU threshold of NMS is 0.5, and the score threshold of the predicted bounding box is 0.05. The max number of the bounding box of each image is set as 100.

\subsection{MS COCO}

\subsubsection{Which fusing strategy is better.}

We propose three different strategies to fusing the features from the image pyramid and the features of the backbone network in this paper. To compare the effectiveness and the difference of them, we perform different strategies in the same baseline and report the $AP$ of small, middle and large objects separately. The results in Table. \ref{tab:fusingstrategy} shows that all three versions have similar results for small objects$(20.8vs18.9vs19)$, but the results for middle objects and large objects have large margin$(2\%-4\%)$ between them. Table. \ref{tab:fusingstrategy} shows that the sum operation achieves much better performance in all metrics. We argue that the sum operation is more suitable for IPG fusion, while product and concatenation are more tricky. Therefore, we perform the rest experiments with a $sum$ fusion module.

\subsubsection{How deep is better for the IPG-Net.}
\begin{table}[h!]
    \centering
    \setlength{\tabcolsep}{0.5mm}{
    \begin{tabular}{c|c|c|c|c|c|c|c}
        \toprule
            \quad model&N stages&$mAP$&$AP50$&$AP75$&$AP_{S}$&$AP_{M}$&$AP_{L}$\\
            \midrule
            IPG RCNN&4&35.4&57.9&37.8&\textbf{21.2}&39.2&44.9\\
            &5&\textbf{35.7}&\textbf{58.2}&{38.2}&\textbf{21.1}&\textbf{39.6}&{45.7}\\
            &6&35.7&58.2&38.3&\textit{20.8}&39.3&45.8\\
            &7(keep)&35.7&58&38.3&\textit{21}&39.6&45.8\\
        \bottomrule
    \end{tabular}}
    \caption{The ablation study of the depth of the backbone of IPG-Net on the COCO \textit{minival}, $7(keep)$ denotes the depth of backbone is 7 stages and the spatial size of those features of the last 3 stages keeps constant.}
    \label{tab:lengthofipgnet}
\end{table}

The Table. \ref{tab:lengthofipgnet} shows that the $mAP$ is not always increasing with the increase of the depth, and we also notice that the improvement comes from the large objects, while the small objects slightly decrease, $0.3\%(21.2vs21.1vs20.8)$. This observation is consistent with the assumption in this paper, shallow layers features are more important for small objects. We also study the effect of keeping the spatial size of the last 3 stages, as the \cite{li2018detnet} proposed. The results show that there is a slight improvement for small objects $(20.8vs21)$ and middle objects $(39.3vs39.6)$, but the performance improvements in $mAP$ is not significant. Considering the computation cost and the model performance, the depth of 5 stages is the best choice for the IPG-Net. Here, we construct the \textbf{IPG RCNN} with a 4 levels IPG-Net and a Faster RCNN head.

\subsubsection{The position of the IPG fusion.}        
\begin{table}[t]
    \normalsize
    \centering
    \setlength{\tabcolsep}{1.5mm}{
    \begin{tabular}{c|c|c|c|c|c|c}
        \toprule
            \quad stage1&stage2&stage3&stage4&$mAP$&$AP50$&$AP75$\\
            \midrule
             -&-&-&-&36.3&58.1&39.0\\
             \cmark&-&-&-&36.5&58.4&39.3\\
             -&\cmark&-&-&36.2&58.1&39.0\\
             -&-&\cmark&-&\textbf{36.6}&58.4&\textbf{39.4}\\
             -&-&-&\cmark&36.5&58.4&39.2\\
             -&\cmark&\cmark&\cmark&36.5&58.4&\textbf{39.4}\\
        \bottomrule
    \end{tabular}}
    \caption{The ablation study of the position of the fusion module in IPG-Net, we add only one fusion module into one stage and also add multi-modules into multi-stages.}
    \label{tab:whereistheipg}
\end{table}
Here we conduct ablation experiments using an IPG-Net and a ResNet with 4 stages. Firstly, we only add one image pyramid feature into the backbone network. Secondly, we increase the level of the image pyramid to find out if more levels are better. The Table. \ref{tab:whereistheipg} shows that IPG-Net with different configures all achieve slight improvement compared with baseline ResNet. The best $mAP$ of them is $36.6\%$, which is only $0.1\%$ improvement from the others. We conclude that the IPG-Net is not sensitive enough for the position of IPG fusion. All in all, IPG-Net indeed improves detection performance. 

\subsubsection{The effect on deep layers.}
\begin{table}[]
    \normalsize
    \centering
    \setlength{\tabcolsep}{1.7mm}{
    \begin{tabular}{c|c|c|c|c|c|c}
        \toprule
            \quad model&$mAP$&$AP50$&$AP75$&$AP_{S}$&$AP_{M}$&$AP_{L}$\\
            \midrule
            IPG-Net&\textbf{23.9}&40.2&\textbf{24.8}&\textbf{4}&\textbf{28.7}&\textbf{39.9}\\
            ResNet&23.6&40.2&24.2&3.9&28.3&39.3\\
            \bottomrule
            \toprule
            \quad model&$AR$&$AR50$&$AR75$&$AR_{S}$&$AR_{M}$&$AR_{L}$\\
            \midrule
            IPG-Net&\textbf{23.7}&\textbf{36.2}&\textbf{38.5}&\textbf{12.4}&\textbf{43.1}&\textbf{60.9}\\
            ResNet&23.4&35.7&38&12&42.3&60.7\\
        \bottomrule
    \end{tabular}}
    \caption{The effect of IPG in deep layers on the COCO \textit{val} based on RetinaNet-50.}
    \label{tab:effectondeeplayer}
\end{table}
As we claimed in this paper, the function of image pyramid guidance is to supply the spatial information and the image details information of small objects into deep features. Here, we conduct a simple comparison experiment to prove the effectiveness of IPG in deep layers. The configuring of the experiment is simple but persuasive. The depth of the IPG-Net and the ResNet are 7 stages but we only use 4 outputs of the last four stages, which are all deep features, without enough detail information as we claimed. The detector we use here is RetinaNet\cite{retina}, whose performance highly relies on the scale of the feature pyramid. 

The Table. \ref{tab:effectondeeplayer} shows that IPG-Net achieves higher performance than ResNet backbone in almost all metrics. The increase of $AP$ reaches $0.6\%(24.8vs24.2,39.9vs39.3)$. The results of Table. \ref{tab:effectondeeplayer} also suggest that the IPG-Net works well as the feature extractor of the RetinaNet\cite{retina}(a one-stage detector). We also notice that the IPG makes more contribution to RetinaNet$(0.6\%)$ than on Faster RCNN $(<0.6\%)$. We argue that's because the two-stage model prefers to perform ROI Pooling in shallow layers' features while the one-stage models consider more deep features.

\subsubsection{Comparison with the state of the art results in MS COCO \textit{test-dev}}
Finally, we also test IPG RCNN in MS COCO \textit{test-dev} to make a comparison with the state of the art detectors. We construct a modified IPG RCNN with an IPG-Net101 and a cascade RCNN head\cite{cai2018cascade}. To reduce the cost and parameters, we choose stage 3 as the level to perform IPG, because the IPG-Net is not sensible with the position of IPG fusion, as mentioned above. The depth of the IPG-Net is four stages to make full use of the pre-trained parameters of standard ResNet in ImageNet. The IPG RCNN achieves $45.7 mAP$ in MS COCO \textit{test-dev}, which is the state of the art result compared with other object detection models under the condition of single scale inference.

\begin{table*}[h]
    \normalsize
    \centering
    \setlength{\tabcolsep}{13mm}{
    \begin{tabular}{c|c|c|c}
        \toprule
            \quad model&backbone&input size&$mAP$\\
            \midrule
            \textit{Two Stage Det}\\
            \hline
            Faster RCNN\cite{resnet}&ResNet-101&~1000x600&76.4\\
            R-FCN\cite{dai2016rfcn}&ResNet-101&~1000x600&80.5\\
            OHEM\cite{OHME}&VGG-16&~1000x600&74.6\\
            HyperNet\cite{hpernet}&VGG-16&~1000x600&76.3\\
            R-FCN w DCN\cite{deformable}&ResNet-101&~1000x600&82.6\\
            CoupleNe\cite{couplenet}t&ResNet-101&~1000x600&82.7\\
            DeNet512(wide)\cite{denet}&ResNet-101&~512x512&77.1\\
            FPN-Reconfig\cite{fpnreconfig}&ResNet-101&~1000x600&82.4\\
            \midrule
            \textit{One Stage Det}\\
            \hline
            SSD512\cite{SSD}&VGG-16&512x512&79.8\\
            YOLOv2\cite{yolov2}&Darknet&544x544&78.6\\
            RefineDet512\cite{refinedet}&VGG-16&512x512&81.8\\
            RFBNet512\cite{RFB}&VGG-16&512x512&82.2\\
            CenterNet\cite{objaspoint}&ResNet-101&512x512&78.7\\
            CenterNet\cite{objaspoint}&DLA\cite{objaspoint}&512x512&80.7\\
            \midrule
            \textit{Ours}\\
            \hline
            Faster RCNN\cite{fasterrcnn}&ResNet-50&~1000x600&79.8\\
            IPG RCNN&IPGnet-50&~1000x600&80.5\\
            IPG RCNN++&IPGnet-50&~1000x600&\textbf{81.6}\\
            IPG RCNN&IPGnet-101&~1000x600&84.8\\
            IPG RCNN++&IPGnet-101&~1000x600&\textbf{85.9}\\
        \bottomrule
    \end{tabular}}
    \caption{The state of the art of the performance on the Pascal VOC 2007 \textit{test}, '++' denotes that inference is performed with three scales.}
    \label{tab:voc}
\end{table*}
\subsection{Pascal VOC}
\subsubsection{Comparison with the state of the art results in Pascal VOC.}
To valid the results more properly, we also test the new \textit{IPG RCNN}(based on Faster RCNN head \cite{fasterrcnn}) in Pascal VOC data set. The baseline is a faster RCNN with the ResNet-50 as a backbone network, the performance of the baseline Faster RCNN is much better than the original paper\cite{fasterrcnn}, reaching $79.8\%mAP$. Then we add the fusion module into \textit{stage 3} following the ablation studies above to construct an IPG RCNN with an IPG-Net50 and a faster RCNN head. The Table. \ref{tab:voc} shows that the IPG-Net-50 obtains $80.5\% mAP$, we further apply multi-scale inference strategy with the resolution $((800,500),(1000,600),(1333,800))$ to test the effort of the IPG-Net-50, resulting in $81.6\% mAP$. Furthermore, to keep consistent with the previous works, we also use a 101 layers IPG-Net to get the state of the art result, the IPG-Net-101 is also fine-tuned with pre-trained parameters on COCO data set. The results of single scale and multi-scale all tested on Pascal VOC2007 \textit{test}. Table \ref{tab:voc} shows that IPG RCNN101 achieves $84.8$ with the single scale test and $85.9$ with the multi-scale test. 

Finally, the results on two popular benchmarks (MS COCO and Pascal VOC) show that the IPG RCNN is robust enough and effective for small/tiny object detection. 




\section{Conclusion}
The main problem we focus on is the information imbalance and misalignment in object detection, especially for small objects. There is a serious information imbalance between the shallow layer and the deep layer for the detection backbone. In this paper, we propose a novel \textit{image pyramid guidance network (IPG-Net)}, including the IPG transformation module and IPG fusion module. The main contribution in this paper is we create a new path to alleviate the imbalance and misalignment problem between the spatial information and the semantic information, fusing the image pyramid information into the backbone stream.
Abundant ablation experiments have been conducted to demonstrate the effectiveness of the IPG-Net. This work also can be extended to the video object detection task further with the natural advantage of the image pyramid guidance. The IPG fusion strategy could also be further investigated, attention-based fusion strategy is a promising path.

\section{Acknowledge}
This work was supported by the National Natural Science Foundation of China under Grant No. 61972036, and in part by Grant No. U1736117 and 61972036.

{\small
\bibliographystyle{ieee_fullname}
\bibliography{egbib}
}

\end{document}